\documentclass[runningheads]{llncs}
\usepackage{graphicx}
\newcommand{\squeeze}{\vspace{-1\baselineskip}}

\begin{document}
\title{The invisible power of fairness. How machine learning shapes democracy}
%
%
\author{{Elena Beretta\inst{1,3}\orcidID{0000-0002-6090-2765} \and
Antonio Santangelo\inst{1}\orcidID{0000-0001-8885-6214} \and
Bruno Lepri\inst{3}\orcidID{0000-0003-1275-2333} \and 
Antonio Vetr\`{o}\inst{1,2}\orcidID{0000-0003-2027-3308} \and
Juan Carlos De Martin\inst{1}}\orcidID{0000-0002-7867-1926}}

\authorrunning{E.Beretta et al.}
%
\institute{Nexa Center for Internet \& Society, DAUIN, Politecnico di Torino, Italy \\
\email{\{elena.beretta,antonio.santangelo,antonio.vetro,demartin\}@polito.it}
\and
Future Urban Legacy Lab, Politecnico di Torino, Italy \and
Fondazione Bruno Kessler, Italy\\
\email{lepri@fbk.eu}}

\maketitle              
\begin{abstract}
Many machine learning systems make extensive use of large amounts of data regarding human behaviors. Several researchers have found various discriminatory practices related to the use of human-related machine learning systems, for example in the field of criminal justice, credit scoring and advertising. Fair machine learning is therefore emerging as a new field of study to mitigate biases that are inadvertently incorporated into algorithms. Data scientists and computer engineers are making various efforts to provide definitions of fairness. In this paper, we provide an overview of the most widespread definitions of fairness in the field of machine learning, arguing that the ideas highlighting each formalization are closely related to different ideas of justice and to different interpretations of democracy embedded in our culture. This work intends to analyze the definitions of fairness that have been proposed to date to interpret the underlying criteria and to relate them to different ideas of democracy. 

\end{abstract}
\squeeze
\squeeze

\section{Introduction}
\label{sec:intro}
\squeeze
Nowadays, machine learning systems make an extensive use of large amounts of data on human behaviors, collected through various channels (e.g. social media, apps, mobile phone usage, credit card transactions, etc.). The widespread resort to these data for disparate purposes is rapidly transforming several domains of our daily life. However, the increasing use of automated-software and machine learning systems, for example in the field of criminal justice \cite{Berk:17}, advertising \cite{Sweeney:13}, and credit scoring \cite{Hardt:16}, is raising a wide range of legal and ethical issues. It is obviously impossible to provide an exhaustive mapping of all the problems, but through an easy expedient a useful logical scheme summarizing the various ones can be supplied. As a matter of fact, if we move from one of the many elementary definitions of ``algorithm" - an encoded procedure to transform \textit{input} (data) in \textit{output} (expected result) through a series of calculations - we understand that we are confronted with three constitutive elements: 1) \textit{input}; 2) \textit{procedure}; 3) \textit{output}. A major concern in the first phase is the presence of biases in the input dataset (this is not the only one: for example, personal data protection is also a relevant issue). In the second phase, a largely debated issue is the transparency and accessibility of the procedure - the problem of the \textit{black box} \cite{Pasquale:15}; while the main problematic aspect of the third phase is the possible discriminatory effects of the algorithmic decision. A given ethical and/or legal problem must be studied with reference to: a) the phase in which it is mainly located, and b) the possible propagation of the issue throughout the algorithm's elaboration path. Since many application domains of machine learning algorithms are not protected by law against discrimination, the attention of actors belonging to different sectors is increasingly focusing on the way algorithms encode prejudices and lead to disproportionate results \cite{ONeal:16}. As a further development of the emergence of adverse outcomes, many researchers are involved in finding solutions to overcome the problem of discrimination in automated software systems by embedding the idea of \textit{fairness} in the algorithm's structure.
In this paper, we contribute to the debate on fairness in machine learning by discussing the impact that its recent mathematical formalizations may have on societies. In particular, we reflect on the meaning of each definition of fairness in relation to the different ideas of democracy and equality they take with them. Our hypothesis is twofold: on one side, we suppose different democratic cultures may foster the diffusion of different definitions of fairness; on the other side, the different ways of designing machine learning tools fit in different ways in the various societies and may fit better than others, especially because in a same social contexts various ideas of democracy may coexist. We describe in Section \ref{sec:fairness} the role of fairness in the field of machine learning; several recent mathematical implementations of fairness and various democracy typologies are exposed and analyzed in Subsection \ref{sec:fairnessML} and Subsection \ref{sec:democracy}, respectively. In Section 3 we discuss commonalities among fairness definitions and democracy typologies, pointing out challenges and open issues for the fair machine learning research domain. Finally, Section 4 draws some conclusions and discuss potential future works on this topic.

\section{Why fairness matters?}
\label{sec:fairness}
\squeeze
The way machine learning systems act is a crucial matter for our societies \cite{ONeal:16}. Algorithms are designed to recognize situations leading to satisfactory outcomes, they are modeled to look for patterns and characteristics in individuals that have historically brought to success, thereby not making things fair if randomly employed, but replicating models and past practices. A relevant example is the recent case, exposed by the investigative website ProPublica\footnote{https://www.propublica.org/article/machine-bias-risk-assessments-in-criminal-sentencing}, about the COMPAS recidivism tool, an algorithm used to inform criminal sentencing decisions by predicting recidivism. In particular, the study found that COMPAS is evaluating useful risk factors for classification, such as socio-economic status and type of crime, which share high covariance with some sensitive attributes, in this case the race. Although the reason for this high covariance may be the result of a preexisting prejudice present in a wider system than one considered for the analysis of recidivism , it should be underlined that the study found that COMPAS was significantly more likely to falsely label as high-risk of recidivism  black defendants because of the above mentioned interaction, and to underestimate the risk of white defendants by less accurately detecting the false negative rate. Although research is increasingly concentrating in finding solutions to mitigate the segregating effect of some algorithms, many efforts will have to be done to consider when an algorithm fails, for whom it fails and what are the social costs of the failure \cite{Corbett:17}. Hence, designing a fair algorithm entails two aspects, closely related to each other: firstly, to evaluate the meaning of choosing one kind of fairness instead of another in a certain society; and secondly, to assess the degree of social acceptability subordinated to the context and to the selected fairness criterion. Considering that more than one definition of fairness cannot be achieved simultaneously \cite{Kleinberg:16}, design choices have a relevant impact on the effect that the algorithm's outcomes will have on society. It is therefore quite relevant starting to figure out which kind of societal values and democratic concepts are tied to the current mathematical formalizations of fairness. In the following sections, we analyze the most relevant mathematical definitions of fairness provided in the machine learning research area, assuming that a decision on how to design a fair machine learning system could be acceptable with more probability as much as it corresponds to the ideas of democracy shaped by individuals in the social context in which it will be used.

\subsection{Fairness in machine learning domain}
\label{sec:fairnessML}
In the recent years, several formal definitions of fairness have been suggested by the machine learning community. In Table \ref{Table:1}, we report the most widespread ones grouped by similar characteristics: in particular, the first column indicates the name of the partitioning, while the second one the extended name of the fairness definition. Finally, the third column contains scientific references which are then further specified in Table \ref{Table:2}.

\begin{table}[ht]

\centering
\begin{tabular}{c|c|c}
\hline
\textbf{Partition} & \textbf{Definition} & \textbf{Reference}\\
\cline{1-3}
Group & Statistical  & [1, 3, 9] \\
        fairness    & parity &  \\
\cline{2-3}
  & Accuracy & [10] \\
            & parity &  \\
\cline{2-3}
 & False positive & [5, 7]  \\
         & parity &  \\
\cline{2-3}
 & Positive rate & [4, 11, 12] \\
        & parity & \\
\cline{2-3}
 & Predictive parity & [6, 7]\\
\cline{2-3}
 & Predictive value parity & [3]\\
\cline{2-3} 
& Equal & [4, 7, 9]\\
& opportunity & \\
\cline{2-3} 
& Equal & [4, 7]\\
& threshold & \\
\cline{2-3} 
& Well-calibration & [2] \\
\cline{2-3} 
& Balance for & [2] \\
& positive class & \\
\cline{2-3} 
& Balance for & [2] \\
& negative class & \\
\cline{1-3} 
Individual &  & [1]\\
fairness &  & \\
\cline{1-3} 
Counterfactual & & [9] \\
fairness & & \\
\cline{1-3} 
Preference-based & Preferred treatment & [8] \\
\cline{2-3} 
fairness & Preferred impact & [8] \\
\cline{1-3} 
Fairness through & & [1, 4, 9] \\ 
unawareness & & \\
\hline
\end{tabular}

\caption{Fairness in machine learning literature}
\label{Table:1}

\end{table}

\begin{table}[ht]
\centering
\begin{tabular}{c|c}
\hline  
                   \textbf{Reference} & \textbf{Paper} \\
                         & \textbf{number} \\
      \cline{1-2}
         \cite{Dwork:12}     &  [1] \\
         \cite{Kleinberg:16} &  [2]   \\
         \cite{Berk:17}      &  [3]   \\
         \cite{Hardt:16}     &  [4]   \\
         \cite{Corbett:17}   &  [5]   \\
         \cite{Simoiu:17}    &  [6]   \\
         \cite{Chouldechova:16}    &  [7]   \\
         \cite{Zafar:17}     &  [8]   \\
         \cite{Kusner:18}    &  [9]   \\
         \cite{Dieterich:16} &  [10]   \\
         \cite{Zafar2:17}    &  [11]   \\
         \cite{Binns:18}     &  [12]   \\
\hline  
\end{tabular}
\caption{References}
\label{Table:2}
\end{table}

First of all, we provide some mathematical notations that compose a typical setup in a machine learning domain:
\textit{X} denotes the features of an individual; 
\textit{Y} denotes the target variable;
\textit{A} denotes a sensitive attribute (i.e. gender, race, etc.);
\textit{C} denotes a classifier;
\textit{S} denotes a score function or a conditional expectation. For example, the frequency of an event given certain observed characteristics can be written as \textit{S} = \textit{E}[\textit{Y}$\mid$\textit{X}];
\textit{t} is a threshold. In case of binary classifiers, the score value causes the acceptance of classifier outputs when it is above \textit{t}, otherwise causes the rejection.

Then, we introduce and briefly describe the fairness definitions listed in Table \ref{Table:1}, supplied with examples regarding risk assessment in the criminal justice domain. Individuals rated high risk of re-offending are classified by 0, otherwise 1 - that means low risk of recidivism. The variable \textit{race} has been considered as a sensitive and protected attribute.

\paragraph{Group fairness.} Below, we introduce several formal definitions of \textit{group fairness}.


\textit{Statistical parity.} Classifier \textit{C} satisfies \textit{statistical parity} if P$_a$(C = 1) = P$_b$(C = 1) for all groups \textit{a, b} - i.e. \textit{a = black, b = white}. This means that both black and white people should have equal probability to be classified as low risk.

\textit{Accuracy parity.} Classifier \textit{C} satisfies \textit{accuracy parity} if P$_a$(C = Y) = P$_b$(C = Y) for all groups \textit{a, b}. This means that both black and white people should have equal probability to be correctly classified as low risk, if belonging to actual low risk rate, and correctly classified as high risk, if belonging to actual high risk rate. 

\textit{False positive parity.} Classifier \textit{C} satisfies \textit{false positive parity} if P$_a$(C = 1$\mid$Y = 0) = P$_b$(C = 1$\mid$Y = 0) for all groups \textit{a, b}. This means that both black and white people with actual high risk rate should have equal probability to be incorrectly classified as low risk (False Positive Rate).

\textit{Positive rate parity.} Classifier \textit{C} satisfies \textit{positive rate parity} if P$_a$(C = 1$\mid$Y = i) = P$_b$(C = 1$\mid$Y = i), \textit{i $\in$ 0, 1}, for all groups \textit{a, b}. This means that both black and white people should have equal probability to be incorrectly classified as low risk - False Positive Rate - and to be correctly classified as low risk (True Positive Rate). 

\textit{Predictive parity.} Classifier \textit{C} satisfies \textit{predictive parity} if P$_a$(Y = 1$\mid$C = 1) = P$_b$(Y = 1$\mid$C = 1), for all groups \textit{a, b}. This means that both black and white people with low risk predicted score (Positive Predictive Value) should have equal probability to really belong to the low risk class.

\textit{Predictive value parity.} Classifier \textit{C} satisfies \textit{predictive value parity} if (P$_a$(Y = 1$\mid$C = 1) = P$_b$(Y = 1$\mid$C = 1)) $\land$ (P$_a$(Y = 0$\mid$C = 0) = P$_b$(Y = 0$\mid$C = 0)) for all groups \textit{a, b}. This means that both black and white people with low risk predicted score (Positive Predicted Value) should have equal probability to really belong to low risk class, and both black and white people with high risk predicted score (Negative Predictive Value) should have equal probability to really belong to high risk class.

\textit{Equal opportunity.} Classifier \textit{C} satisfies \textit{equal opportunity} if P$_a$(C = 1$\mid$Y = 1) = P$_b$(C = 1$\mid$Y = 1) for all groups \textit{a, b}. This means that both black and white people with actual low risk rate should have equal probability to be incorrectly classified as high risk (False Negative Rate). Since mathematically a classifier that satisfies False Negative Rate equity satisfies at the same time True Positive Rate equity, the definition also implies that both black and white people with actual low risk rate should have equal probability to be correctly classified as low risk.

\textit{Equal threshold.} Classifier \textit{C} satisfies \textit{equal threshold} if P$_a$(Y = 1$\mid$S = s) = P$_b$(Y = 1$\mid$S = s), \textit{s $\in$ [0, 1]}, for all groups \textit{a, b}. 
This means that both black and white people should have equal score threshold \textit{t} under which they are classified at low risk, and above which they are classified at high risk. 

\textit{Well-calibration.} Classifier \textit{C} satisfies \textit{well-calibration} if P$_a$(Y = 1$\mid$S = s) = P$_b$(Y = 1$\mid$S = s) = s, \textit{s $\in$ [0, 1]}, for all groups \textit{a, b}. 
This means that both black and white people with the same score should be treated comparably ``\textit{with respect to the outcome, rather than treating black and white people with the same score differently based on the race group they belong to}"\cite{Kleinberg:16}.

\textit{Balance for positive class.} Classifier \textit{C} satisfies \textit{balance for positive class} if E$_a$(S$\mid$Y = 1) = E$_b$(S$\mid$Y = 1), for all groups \textit{a, b}. This means that both black and white people with an actual low risk rate should have the same expected value assigned by the classifier \textit{C} (a classifier uses the characteristics of individuals to identify which class - or group - they belong to). That is to say, it should not happen that the scoring process is ``systematically more inaccurate for negative cases - high risk score - in one group than the other"\cite{Kleinberg:16}.

\textit{Balance for negative class.} Classifier \textit{C} satisfies \textit{balance for negative class} if E$_a$(S$\mid$Y = 0) = E$_b$(S$\mid$Y = 0), for all groups \textit{a, b}. This means that both black and white people with an actual high risk rate should have the same expected value assigned by the classifier \textit{C}.
That is to say, it should not be that the scoring process is "systematically more inaccurate for positive cases - low risk score - in one group than the other"\cite{Kleinberg:16}.

\paragraph{Individual fairness.} Given a set of individuals \textit{V} and a set of outcomes \textit{A = $\left \{ 0, 1 \right \}$}, and considering a metric on individuals \textit{d: V x V $\to$ R} and randomized mappings \textit{M: V $\to$ $\Delta$A}, \textit{individuals fairness} is achieved if a randomized classifier, mapping individuals to distributions over outcomes, minimizes expected loss subject to the (D,\textit{d})-Lipschitz condition of \textit{D(Mx, My) $\le$ d(x, Y)} \cite{Dwork:12}. \\
This means that two individuals are similarly classified if they are considered similar with respect to a particular task, such as to pay off a debt with a bank.

\paragraph{Counterfactual fairness.} Classifier \textit{C} satisfies \textit{counterfactual fairness} if P(C$_A$$_\gets$$_a$ (U\footnote{"\textit{U is a set of latent background variables, which are factors not caused by any variable in the set V of observable variables}"\cite{Kusner:18}}) = y$\mid$X = x, A = a) = P(C$_A$$_\gets$$_{a'}$ (U) = y$\mid$X = x, A = a). 
That is, given a set of attributes (\textit{education level, type of crime, drugs problems} and protected attribute \textit{A = race}) and an outcome $\hat{Y}$ to be predicted (\textit{recidivism}), a graph is counterfactually fair if \textit{race} is not directly linked to $\hat{Y}$ through any other attributes. 

Intuitively, this means that a decision is fair towards an individual if it is the same in (i) the actual world and (ii) a counterfactual world where the individual belonged to a different demographic group (i.e. white instead of black).

\paragraph{Preference-based fairness.} Here, we introduce new formalization of fairness \cite{Zafar:17} that are inspired by the concepts of fair division in economics and game theory \cite{varian1974,berliant1992}.

\textit{Preferred treatment.} Classifier \textit{C} satisfies \textit{preferred treatment} if B$_a$(C$_a$) $\ge$ B$_{a'}$(C$_{a'}$), for all a, a' $\in$ A\footnote{\textit{B$_a$ is the fraction of beneficial outcomes received by users sharing a certain value of the sensitive attribute a}\cite{Zafar:17}}. This means that the preferred condition is preserved if each group obtains more benefit from their own classifier then it would be assigned from any other classifier. 
In other words, both black and white people should prefer \textit{``the set of decisions they receive over the set of decisions they would have received had they collectively presented themselves to the system as members of a different sensitive group"}\cite{Zafar:17}.

\textit{Preferred impact.} Classifier \textit{C} satisfies \textit{preferred impact} if B$_a$(C) $\ge$ B$_a$(C'), for all a $\in$ A. This means that the preferred condition is preserved if a classifier C, with respect to any other classifier, assigns at least the same benefit for all groups. In other words, both black and white people should prefer \textit{``the set of decisions they receive over the set of decisions they would have received under the criterion of impact parity"}\cite{Zafar:17}.

\paragraph{Fairness through unawareness.} Classifier \textit{C} satisfies \textit{fairness through unawareness} if X: X$_a$ = X$_b$ $\to$ C$_a$ = C$_b$ for both individuals \textit{a, b}. This means that for example the attribute \textit{race} should not be used to train the classifier and thus to take a decision (i.e. granting or not a loan). 

\subsection{Fairness and democracy}
\label{sec:democracy}
In order to understand which idea of fairness can be affirmed in a certain society, it may be useful to relate it to the type of democracy in force within the latter. In this way, we may reflect on the concept of equality that this idea of democratic life brings with it. As the philosopher Christiano writes \cite{Christiano:06}, democracy is a procedure for taking collective decisions characterized by the fundamental equality of those who participate in it. According to Bozdag and Van Den Hoven \cite{Bozdag:15}, there are at least four different ways of conceiving democracy, which must be seen as some sort of ideal types, that interact and confront each other in reality, giving rise to the different democratic regimes in force in the world. The first is the model of \textit{liberal democracy}, commonly centred on the defence of fundamental rights and freedoms (i.e. self-determination, choice, private property, expression, etc.) from all forms of coercive power \cite{Dunn:79}, \cite{Held:06}. It is an individualistic vision of democratic life, according to which democracy must be the place that preserves the freedoms of the individual, characterizing itself as a procedure which, through electoral competition and voting, represents and aggregates the will of individuals. In this case, the criterion of justice on which the concept of equality between persons is based denies the maxim of egalitarianism, according to which all men must be (at the very least) equal in everything, but admits the equality of all only in something, that is, in the so-called fundamental rights \cite[p.~95]{Bobbio:95}. The second model is that of \textit{deliberative democracy} \cite{Bozdag:15}, seen as a way of taking collective decisions, based on a confrontation between free and equal citizens. This confrontation must be aimed at finding logical solutions to satisfy the good of all. Here, the emphasis is less on the individual and more on her/his participation in common life. Democracy is not seen as the mere aggregation of the will of individuals in a vote, but as a place where people can talk to each other, make their points of view count and try to understand, together, which is the best one. The fulcrum of all this is the exercise of a rationality, in a certain sense, devoid of the will to affirm partisan interests, which is very similar to the rationality theorized by Habermas \cite{Habermas:98} and Rawls \cite{Rawls:71}, \cite{Rawls:97}. In this case, equality is above all in the possibility of having access to the deliberative process and in ensuring that one can confront others, even if s/he is part of a minority. The third model of democracy is what Bozdag and Van Den Hoven \cite{Bozdag:15} call \textit{republican} or \textit{protesting}. It is based on the idea that all citizens must be free from the arbitrary exercise of power by someone. For this reason, they must be able to control the activity of those who govern them, being at the same time able to challenge it. More than the consensus of deliberative democracy, therefore, in this case it is fundamental the power to challenge choices and points of view that are not shared. It is very important the transparency and publicity of the actions of the rulers, and it is essential that anyone can question them, in a way that can produce some effect. This is the kind of equality that the supporters of this model of democracy are aiming for. The fourth model is that of \textit{competitive democracy} \cite{Mouffe:09}. It logically contrasts with that of deliberative democracy, because its proponents do not believe that people, by confronting and using pure rationality, can reach an agreement on the common good. Often, the positions of each one are based on passions, partisan interests, instances that one does not want or cannot negotiate. Democracy, then, must consist of a set of procedures that allow the supporters of the various positions to express themselves and to confront each other, in order to prevail. This must be done knowing that the opponents, having lost a confrontation, will have the opportunity to assert their demands in future occasions. Equality, in this case, is similar to the one existing in a competitive context, in which the participants have, at the beginning, the same opportunities to assert themselves. It is defined, in fact, as \textit{equality of opportunity} \cite[p.~24-26]{Bobbio:95}. To these four models, however, it is necessary to add a fifth that we could define as \textit{egalitarian democracy}, following Bobbio's reasoning \cite[p.~26-38]{Bobbio:95}. It is based on the ideal of guaranteeing a \textit{de facto} equality which, as Bobbio himself writes (ibidem, p. 27), is economic equality. In practice, egalitarian democracy tries to reduce the distance between those who have less and those who have more, with equalizing mechanisms such as progressive taxes, subsidies to the poor, and so on. Hence, the principle on which it is based is the opposite of that of competitive democracy, in the sense that its proponents recognize the differences between people rather than what they have in common, and try to favor those who are disadvantaged. 

\section{Results}
\label{sec:results}
\begin{figure}[ht]
    \centering
    \includegraphics[width=0.8\textwidth]{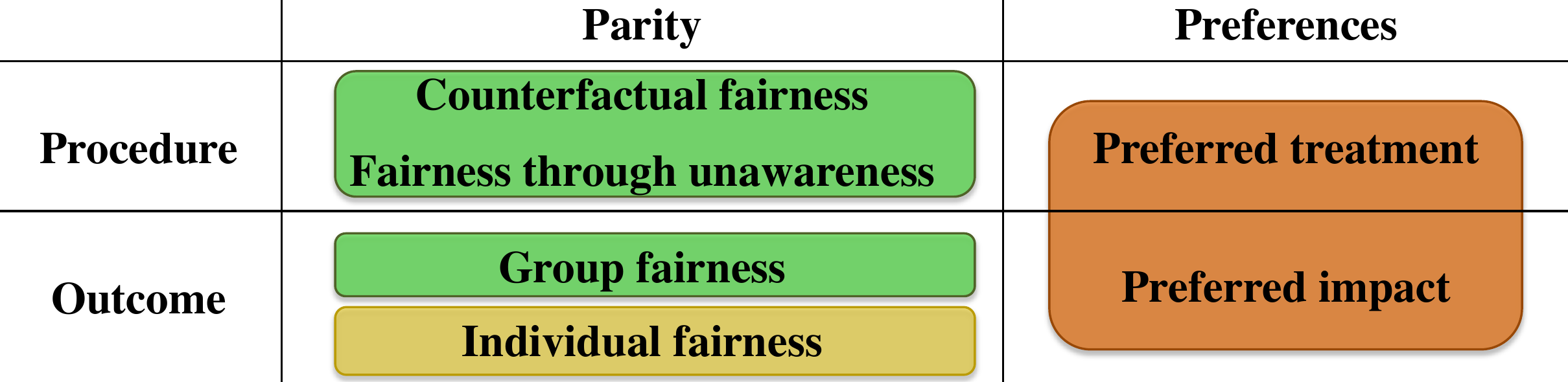}
    \caption{Trade-off in fairness selection process related to democracy typologies.
    Legend: \textit{Competitive} (green), \textit{Liberal} (yellow), \textit{Egalitarian} (orange)}
    \label{fig:1}
\end{figure}

\squeeze
In the previous section we exposed the most widespread fairness criteria formalized in the machine learning domain, observing that the achievement of fairness is subordinated to various probabilistic constraints, each of them related to one or more parts which an algorithm is composed of (see Section \ref{sec:intro}). The connection among machine learning approaches and democratic structures similarly exists: as well as fair parameters can be encoded both in procedure and in algorithms' outcome, even democratic values can be accomplished both in procedural and in outcome part of a decision. Therefore, as we show in Figure \ref{fig:1}, selecting a fairness criterion instead of another is a trade-off process where the decision-maker is called upon to determine if the achievement of the desired goal entails preserving parity (all individuals are equal on the basis of some protected attributes) or satisfying preferences (individuals may have different preferences independent from protected attributes), and if it has to be reached by means of procedure constraints or impact factors. In other words, should the algorithm be fair in the way it takes a decision or should the decision be fair itself? The first column refers to the practice of preserving a kind of equality among people, while the second one refers to the practice of satisfying preferences; rows indicate in which part of an algorithm the fairness criteria are embedded and if inequality is mitigated from the beginning or in a second phase. The following partitioning (Figure \ref{fig:1}), inspired by Gajane and Pechenizkiy \cite{Gajane:18} is the result of our reasoning, which is explained in details for each democracy type in the following subsections.

\paragraph{Liberal democracy} Liberal democracy is a model based on self-determination and preservation of individual freedom. Analyzing the mathematical formulation of \textit{individual fairness}, we believe that such a notion of fairness underlies an idea of liberal democracy; in fact, even though attention is paid to protect minorities' rights, individual diversity is chiefly preserved in preferential way, considering individuals not because of an intrinsic differentiation but rather because of some different tasks for which they are similarly treated. 
\paragraph{Competitive democracy} 
This kind of democracy expresses the idea that actors cannot reach an agreement among themselves on a specific result by reason of utility; this reasoning implies the condition occurring when allocation of goods is such that it is not possible to improve the condition of one actor without worsening the condition of another one. So the best that can be done is guaranteeing equality of opportunity among groups through preserving and protecting minorities' rights. Therefore, from the point of view of competitive democracy, the mechanism of satisfying preferences could be seen as considering like a game which each individual should have the same opportunity to play. In this sense, almost all the fairness definitions that we have reported reflect the competitive democracy criteria; in fact, the majority of them encodes the idea that all people should have the same possibility of being correctly classified (i.e. each individual should have the same opportunity to receive a loan if s/he deserves it). From a statistically point of view, fairness definitions clustered in the \textit{group fairness} partition are not substantially very different, the disparity mainly consists on the observer's point of view, or rather the stakeholder's; i.e., in \textit{statistical parity} we assume the society's point of view that wonder \textit{"Is the selected set demographically balanced?"}; in \textit{false positive parity} we assume the defendant's point of view that wonder \textit{"What is the probability that I'm incorrectly classified as high-risk?"}; in \textit{predictive value parity} we assume the decision-maker point of view that wonder \textit{"Of those I have labeled at high risk, how many will be recidivists?"}.
\paragraph{Egalitarian democracy} This type of democracy is inspired to Bobbio's notion of \textit{distributive egalitarianism}, intended as a way of redistributing material resources between advantaged and disadvantaged people \cite[p.~30-38]{Bobbio:95}. But if we extend the idea of redistribution also to political resources, knowledge, etc., then we can apply it to the fairness criteria we have reported. In fact, no definition of fairness completely fulfills egalitarian democracy criteria, but preference-based models are the closest ones; in fact, both definitions of preference-fairness refer to the concept of \textit{benefit}, which as expressed in the definition of Zafar \emph{et al.} \cite{Zafar:17} is very similar to the concept of favoring individuals belonging to minorities.
We exclude republican and deliberative democracy from the general classification reported in Figure \ref{fig:1}.
The problems that are faced, when reflecting on the republican model of democracy, are usually touched, within the ethics of data and algorithms, when referring to the issues of transparency and accountability of computer tools. Since the attention mainly focuses on monitoring who exercises the power and on controlling the decisions, we believe actually the whole general ground of fair machine learning has been inspired by \textit{republican democracy} values. But none of the definitions of fairness we have reported have directly to do with transparency and accountability, which have to be considered as the results of the whole fairness process. \\

Instead, \textit{deliberative democracy} is a model based on agreement and on the idea that all individuals are free and equal to each others, when having the possibility of seeing their instances represented and taken into consideration in the contexts where decisions about them and their lives are taken. Its ethical premises are very similar to the ones of \textit{competitive democracy}. In fact, strong attention is given to preserve minorities' rights and to give everyone equal possibilities to participate to the game of democratic debate. But once again, we think none of the fairness definitions we have reported has directly to do with guaranteeing that democratic decisions are taken acknowledging the instances of everybody. Moreover, we believe at the present the field of fair machine learning still lacks fairness definitions built on egalitarian democratic principles. In fact, if almost all fairness definitions are to some extent based on the principle that everyone should have equality of opportunities, egalitarian democratic definitions can be considered in contrast to this principle, because they rely on a sort of an \textit{inequality principle}, according to which a disadvantaged individual must be favored over others. Criteria relating to most of the fairness definitions are linked to competitive democracy which may be a following stage of the egalitarian democracy, because competitive democracy rules, according to which everyone should have the opportunity to fairly participate at the competition, can be better implemented if inequalities are previously resolved through a distributive logic. 
\section{Discussion and Conclusions}
\squeeze
Over the last decade machine learning has radically changed the way we take a decision. Many researchers studied the performances of statistical models and human judgments, demonstrating how in some circumstances models based on machine learning have surpassed those based on human judgments (\cite{Barocas:18}). 
Moreover, by now machine learning algorithms are well known to use a large-amount of data to induce a particular rule starting from a generalized datum; in this article we have addressed problems related to the correctness of these rules. The examples enhancing machine learning models have in fact shown how often the results reflect bias and human prejudices in different contexts - especially those in which we try to reproduce and analyze human behavior. Hence, several efforts have been devoted to finding solutions that re-calibrate balances in outputs of machine learning systems. 
However, the debate around \textit{fairness} the in machine learning domain displays a profound and relatively worrying lack: although researchers are acting and reacting in a positive and proactive way, data scientists and computer engineers are increasingly involved in taking decisions that affect individuals, operating as judges in decision-making processes and constituting a sort of \textit{invisible power} - although in a positive way - whereas society at large should play this role. In this article, we enrich these efforts and the related debate, highlighting that meanings of fairness, underlying to statistical constraints, and the democratic values are strictly connected; in fact, we carry on the idea that the spread of one fairness definition instead of another is highly justified by the ideas of justice and democracy shared among the society in which the above-mentioned fairness criterion is selected. We consider this qualitative analysis as a prelude of quantitative researches. For example, our study may pave the way for the implementation of an agent-based model that mimics the decision-making, where agents are informed by different kind of democratic values. The aim of the model would be to analyze which statistical parameters lead to specific fairness criteria and under which probabilities. Moreover, we plan to further develop our work by proposing a new formalization of fairness based on egalitarian democratic principles.

%
%
%
\bibliographystyle{splncs04}

\begin{thebibliography}{9}

\bibitem{Barocas:18}
  Barocas,  S., Hardt,  M., Narayanan,  A., Fairness  and  Machine  Learning. fairml-book.org (2018), \url{http://www.fairmlbook.org}

\bibitem{Berk:17}
  Berk, R., Heidari, H., Jabbari, S., Kearns, M., Roth, A., Fairness in criminal justice risk assessments: The state of the art. Sociological Methods {\&} Research (2018)

\bibitem{berliant1992}
  Berliant, M., Thomson, W., On the fair division of a heterogeneous commodity. Journal of Mathematics Economics (1992)
  
\bibitem{Binns:18}
  Binns,  R., Fairness  in  machine  learning:  Lessons  from  political  philosophy.  In:Proceedings of Machine Learning Research. vol. 81, pp. 149–159. Sorelle A. Friedler, Christo Wilsonf (2018)

\bibitem{Bobbio:95}
  Bobbio, N., Eguaglianza e libert{\'a}. Einaudi, Torino, Italy (1995)
  
\bibitem{Bozdag:15}
  Bozdag, E., van den Hoven, J., Breaking the filter bubble: democracy and design.Ethics and Information Technology17, 249–265 (2015) 

\bibitem{Chouldechova:16}
  Chouldechova, A., Fair prediction with disparate impact: A study of bias in recidivism prediction instruments. Big Data (2017)

\bibitem{Christiano:06}
  Christiano, T., Democracy. Stanford Encyclopedia of Philosophy (2006)

\bibitem{Corbett:17}
  Corbett-Davies, S., Pierson, E., Feller, A., Goel, S., Huq,  A., Algorithmic  decision making and the cost of fairness. In: Proceedings of the 23rd ACM SIGKDD International Conference on Knowledge Discovery and Data Mining (KDD 2017)(2017)

\bibitem{Dieterich:16}
  Dieterich, W., Mendoza, C., Brennan, T.,  Compas risk scales: Demonstrating ac-curacy equity and predictive parity. Tech. rep., Northpointe Inc. (2016)

\bibitem{Dunn:79}
  Dunn,  J., Western political theory in the face of the future. Vol. 3. Cambridge University Press, Cambridge, UK (1979)

\bibitem{Dwork:12}
  Dwork,  C., Hardt, M., Pitassi, T., Reingold, O., Zemeln, R., Fairness through awareness. In: Proceedings of the 3rd Innovations in Theoretical Computer Science Conference. pp. 214–226. ACM (2012)

\bibitem{Gajane:18}
  Gajane, P., Pechenizkiy, M., On formalizing fairness in prediction with machine learning. arXiv:1710.03184 (2018)

\bibitem{Habermas:98}
  Habermas, J., Between facts and norms: Contributions to a discourse theory of law and democracy. MIT Press, Cambridge, US (1998)

\bibitem{Hardt:16}
  Hardt, M., Price, E., Srebro, N., Equality of opportunity in supervised learning. In: Advances in Neural Information Processing Systems (2016)
  
\bibitem{Held:06}
  Held, D., Models of democracy. Stanford University Press, Palo Alto, US (2006)

\bibitem{Mouffe:09}
  Held, D., The democratic paradox. Verso, London, UK (2009)

\bibitem{Kleinberg:16}
  Kleinberg, J., Mullainathan, S., Raghavan, M., Inherent trade-offs in the fair determination of risk scores. In: Proceedings of Innovations in Theoretical Computer Science (ITCS 2017) (2017)

\bibitem{Kusner:18}
  Kusner, M.J., Loftus, J.R., Russell, C., Silva, R., Counterfactual fairness. In: Proceedings of 31st Neural Information Processing Systems (NIPS 2017) (2017)
  
\bibitem{ONeal:16}
  O’Neil, C., Weapons of Math Destruction: How Big Data Increases Inequality and Threatens Democracy. Crown Publishing Group, New York, NY (2016)

\bibitem{Pasquale:15}
  Pasquale, F., The Black Box Society: The Secret Algorithms That Control Money and Information. Harvard University Press, Cambridge, MA, USA (2015)
  
\bibitem{Rawls:71}
  Rawls, J., A theory of justice. Harvard University Press, Harvard, US (1971)

\bibitem{Rawls:97}
  Rawls, J., The idea of public reason. The MIT Press, Cambridge, US (1997)

\bibitem{Simoiu:17}
  Simoiu, C., Corbett-Davies, S., Goel, S., The problem of infra-marginality in out-come tests for discrimination. The Annals of Applied Statistics (2017)
  
\bibitem{Sweeney:13}
  Sweeney, L., Discrimination in online ad delivery. Queue - Storage11(3) (2013)

\bibitem{varian1974}
  Varian, H.: Equity, envy, and efficiency. Journal of Economic Theory (1974)

\bibitem{Zafar2:17}
  Zafar, M.B., Valera, I., Rodriguez, M.G., Gummadi, K.P., Fairness  beyond  disparate treatment and disparate impact: Learning classification without disparate mistreatment. In: Proceedings of the 26th International Conference on World Wide Web (WWW 2017) (2017)
  
\bibitem{Zafar:17}
  Zafar, M.B., Valera, I., Rodriguez, M.G., Gummadi, K.P., Weller, A.: From parity to preference-based notions of fairness in classification. In: Proceedings of the 31st Conference on Neural Information Processing Systems (NIPS 2017) (2017)

\end{thebibliography}

%

\end{document}